\documentclass[11pt,bezier,amstex]{article}

\usepackage[in]{fullpage}
\usepackage{authblk}
\usepackage[utf8]{inputenc} 
\usepackage[T1]{fontenc}    
\usepackage{times}
\usepackage{soul}
\usepackage{url}            
\usepackage[hidelinks]{hyperref}
\usepackage[utf8]{inputenc}
\usepackage[small]{caption}
\usepackage{graphicx}
\usepackage{amsmath}
\usepackage{amsthm}
\usepackage{amsfonts}       
\usepackage{nicefrac}       
\usepackage{microtype}      

\usepackage{booktabs}       
\usepackage{algorithm}
\usepackage{algorithmic}
\usepackage[switch]{lineno}

\usepackage{xcolor}         
\usepackage{natbib}

\usepackage{graphicx}
\usepackage{subfigure}
\usepackage{caption}

\usepackage{makecell}
\usepackage{multirow}
\usepackage{multicol}

\usepackage{amsmath,enumerate}
\usepackage{amssymb,amsopn,algorithm,algorithmic,float,bbm,bm,enumerate,color,multirow,gensymb}
\usepackage{thmtools, thm-restate}
\usepackage{mathtools}

\usepackage{tabularx}
\usepackage{hhline}

\newcommand{\name}{DeCom}
\linenumbers
\newtheorem{remark}{Remark}

\title{\name: Deep Coupled-Factorization Machine for Post COVID-19 Respiratory Syncytial Virus Prediction with Nonpharmaceutical Interventions Awareness}

\author{Xinyan~Li}
\author{Cheng~Qian}
\author{Lucas~Glass}
\affil{Analytics Center of Excellence, IQVIA Inc., United States}

\begin{document}
\nolinenumbers
\maketitle

\let\thefootnote\relax\footnotetext{Corresponding Author: Xinyan Li (xinyan.li3@iqvia.com).}

\begin{abstract}
Respiratory syncytial virus (RSV) is one of the most dangerous respiratory diseases for infants and young children. Due to the nonpharmaceutical intervention (NPI) imposed in the COVID-19 outbreak, the seasonal transmission pattern of RSV has been discontinued in 2020 and then shifted months ahead in 2021 in the northern hemisphere. It is critical to understand how COVID-19 impacts RSV and build predictive algorithms to forecast the timing and intensity of RSV reemergence in post-COVID-19 seasons. In this paper, we propose a deep coupled tensor factorization machine, dubbed as DeCom, for post COVID-19 RSV prediction. DeCom leverages tensor factorization and residual modeling. It enables us to learn the disrupted RSV transmission reliably under COVID-19 by taking both the regular seasonal RSV transmission pattern and the NPI into consideration. Experimental results on a real RSV dataset show that DeCom is more accurate than the state-of-the-art RSV prediction algorithms and achieves up to 46\% lower root mean square error and 49\% lower mean absolute error for country-level prediction compared to the baselines.
\end{abstract}

\section{Introduction}
\label{sec:intro}


Respiratory syncytial virus (RSV) is the most common cause of bronchiolitis and pneumonia in infants, young children, and seniors \citep{hall2009burden}. According to the Centers for Disease Control and Prevention (CDC), about 58,000 children under the age of five are hospitalized in the United States every year as a result of RSV infection \citep{cdcrsv}. Worldwide, this number can reach more than 3 million every year \citep{shi2017global}, and the true burden is likely to be much more, with approximately half of RSV-associated deaths estimated to occur outside of hospitals \citep{shi2017global}.

Prior to the 2020–21 season, RSV has had consistent and predictable seasonal epidemics in both temperate and tropical regions. Such epidemics often begin in the tropics of each hemisphere in the late summer and reach most temperate locations in the winter and early spring. For instance, in the United States, RSV activity typically begins in Florida in the fall and gradually spreads from the southeast to the northwest. However, nonpharmaceutical interventions (NPI), such as physical separation, stay-at-home directives, school closings, travel bans, and border closures have been put in place as a result of the current COVID-19 outbreak.
While NPI minimizes COVID-19 transmission, it has also had substantial effects on the circulation of RSV.

Since the implementation of NPIs, which are utilized to contain the COVID-19 pandemic, many regions have experienced uncertainty over the RSV outbreaks. For instance, significant declines in RSV infection have been observed globally in 2020. Even during the RSV seasons, when the number of infectious cases was often consistently high, this number was close to zero in many countries. However, nothing has yet returned to normal in 2021, when most countries experienced out-of-season RSV outbreaks that arrived months ahead of the normal RSV season. For example, in the northern hemisphere, France, Spain, and the United States have observed the RSV waves in the spring and summer, similar to countries in the southern hemisphere such as Australia and South Africa, which have also witnessed abnormal out-of-season RSV waves. Until now, we are still suffering from an abnormal RSV epidemic. Therefore, comprehending the association of different factors with the timing and intensity of re-emergent RSV epidemics is crucial for clinical and public health decision-making in post COVID-19 seasons.

Predicting post COVID-19 RSV transmission with NPI-awareness is a challenge. Limited research has achieved satisfactory results for this task. 
In the literature, the most related work is \citet{baker2022}, where the authors employed a time-series Susceptible–Infected–Recovered (TSIR) model to simulate future trajectories of RSV. This approach is based on the standard SIR model and it assumes that all susceptible cases will be infected once. Its limitation is that the number of susceptible people and the infectious rate are difficult to estimate with COVID-19 NPIs, and it also fails to model the COVID-19 impact on RSV transmission. Therefore, TSIR works more like an analytical tool than a predictive model. 

In this paper, we propose a new deep coupled tensor factorization approach named \name~for post COVID-19 RSV prediction with NPI awareness. \name~consists of two coupled spatiotemporal tensor factorization units which give us the ability to model the regular seasonal pattern of the RSV and understand how COVID-19-induced NPIs affect RSV transmission. In order to demonstrate the efficacy and high accuracy of \name, we provide experimental results utilizing real RSV datasets and compare them to state-of-the-art machine learning algorithms.


\subsection{Organization of the paper.} 
We begin with a review of related work in Section \ref{sec:related_work}. Section \ref{sec:prelim} provides a brief overview of Canonical Polyadic Decomposition (CPD), a crucial element of our proposed \name~model. After that, in Section \ref{sec:problem}, we give a formal description of the specific RSV forecasting problem. Section \ref{sec:model} then goes into depth about the proposed \name~model. The data and experimental setting are then described in Section \ref{sec:setup}, and Section \ref{sec:result} discusses the experimental results, comparing the predictive skill of \name~with a number of benchmark models. Finally, Section \ref{sec:conclusion} draws the conclusion.

\section{Related Work}
\label{sec:related_work}

\textbf{RSV prediction and environmental drivers.} \citet{baker2019epidemic,pitzer2015environmental} have shown that RSV epidemics exhibit limit cycle behavior tuned by climate-driven seasonality. In most regions in the United States, RSV and influenza exhibit peak incidence in the winter months when the weather is cold and dry \citep{baker2019epidemic,shaman2010absolute}. There have been attempts to link climatic variables with RSV seasonality \citep{paiva2012shift,alonso2012comparative,du2009meteorological,lapena2005climatic,noyola2008effect,walton2010predicting,yusuf2007relationship}. For example, \citet{pitzer2015environmental} employed linear models to connect incidence patterns to candidate environmental drivers of RSV epidemics in US states. \citet{walton2010predicting} have constructed Na\"{i}ve Bayes (NB) classifier models to predict RSV outbreaks in Salt Lake County, Utah. The given NB models used weather data from 1985 to 2008, considering only variables that are available in real time and can make forecasts up to 3 weeks in advance with an estimated sensitivity of up to 67\% and estimated specificities as high as 94\% to 100\%. However, \citep{white2005transmission, white2007understanding, leecaster2011modeling,weber2001modeling} have found RSV epidemics show strong signatures of nonlinear epidemic dynamics. Few studies have explored climatic associations in a variety of locations covering a range of RSV seasonality patterns and climatic regimes \citep{yusuf2007relationship}. \citep{altizer2006seasonality} refined \citet{pitzer2015environmental}'s linear models to account for the dynamics of infection and fluctuations in immunity and susceptibility that can influence the relationship between environmental factors and epidemic timing .


\subsection{Transmission dynamics of RSV.} 
Mathematical models have been used in epidemiology to characterize annual epidemics of respiratory infections and estimate the likely effectiveness of intervention strategies.  
Using ordinary differential equations (ODEs), many models have been developed in the past to describe the transmission dynamics of RSV \citep{white2005transmission, white2007understanding, leecaster2011modeling,weber2001modeling,paynter2014using,arenas2009stochastic}. The Susceptible-Infected-Recovered (SIR) model \citep{kermack1927contribution}, the Susceptible-Exposed- Infected-Recovered (SEIR) \citep{cooke1996analysis} and their variants, such as SIS, SIRS, and delayed SIR, are only a few examples of such models. For instance, \citet{acedo2010mathematical} introduced an age-structured SIRS model, while \citet{moore2014modelling} developed a seasonally forced compartmental age-structured Susceptible-Exposed-Infectious-Recovered-Susceptible (SEIRS) model to fit to the seasonal curves of positive RSV detections using the Nelder-Mead method.

Most recently, \citet{baker2019epidemic} used a time-series Susceptible-Infected-Recovered (TSIR) model to estimate future trajectories of RSV. The simple SIR component of their model assumes that each susceptible case will contract the infection once. Therefore, this does not represent the actual situation in which susceptible cases transform into exposed cases first, and then a proportion of exposed cases become infected.

\subsection{ML based RSV predictive models.} Modern machine learning approaches have also shown to be beneficial in addition to mathematical models. 
\citet{reis2019superensemble} proposed a superensemble approach to combine three forecasting methods, for RSV prediction in the US at three different spatial resolutions: national, regional, and state. They used three methods to generate retrospective forecasts of RSV outbreaks, including: 1) Bayesian Weighted Outbreaks (BWO), 2) the ensemble adjusted Kalman filter (EAKF) combined with a dynamical susceptible-infected-recovered (SIR) model, and 3) the null model using historical expectance \citep{reis2016retrospective}. \citet{gebremedhin2022developing} fitted a multivariable logistic regression model to identify predictors of RSV-positivity (binary outcome) amongst children under the age of five who were hospitalized and tested for RSV in Western Australia.

\begin{figure*}[t]
\centering
\includegraphics[width=0.98\textwidth]{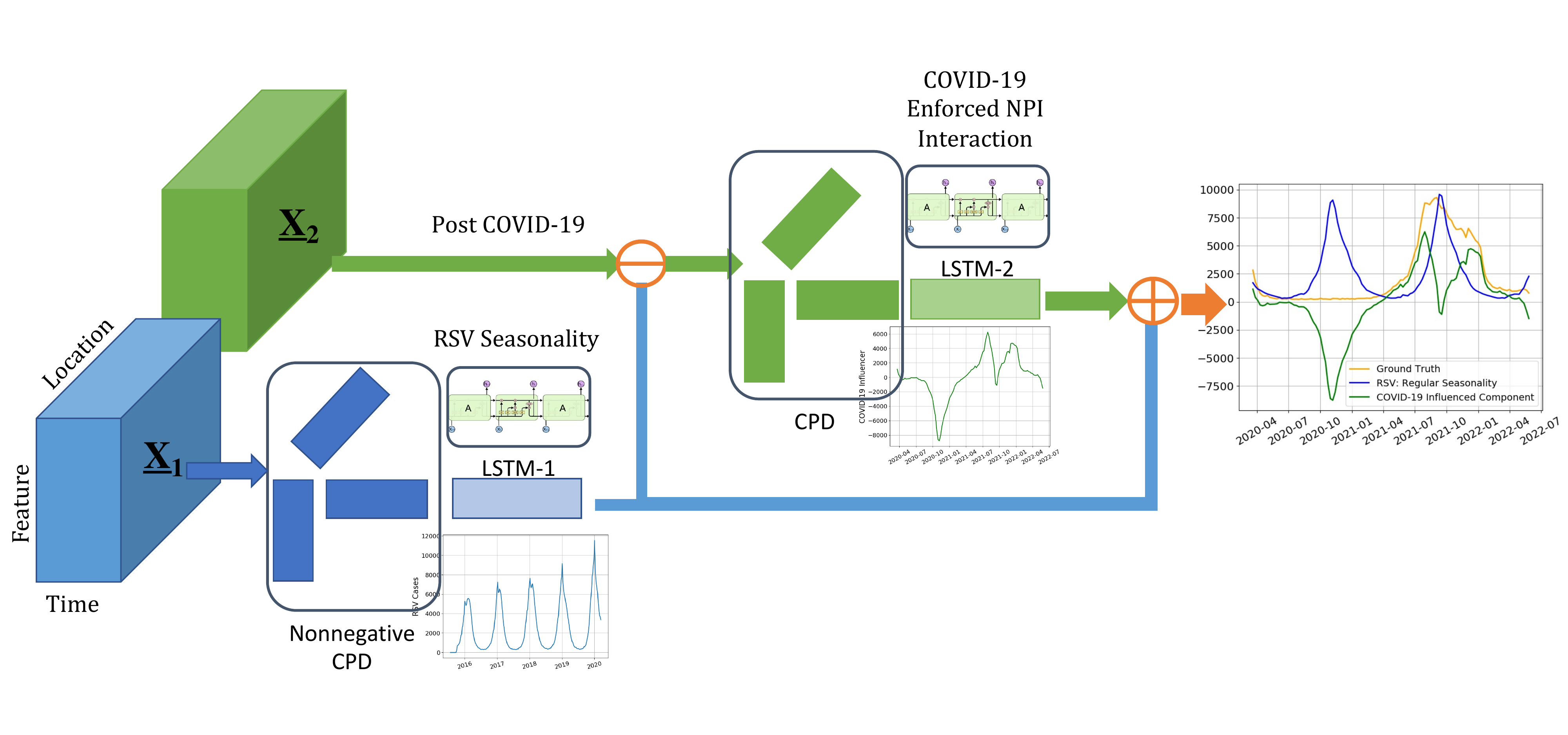}
\caption{Architectures of the \name~model. 
}
\label{fig:decom}
\end{figure*}

\section{Preliminaries}
\label{sec:prelim}

In this section, we review the Canonical Polyadic Decomposition (CPD), an essential step for developing our model. The notation used throughout the paper has been summarized in Table~\ref{tab:notation}.

\subsection{Canonical Polyadic Decomposition.} 
The Canonical Polyadic Decomposition (CPD) approximates a 3-way tensor $\underline{\mathbf{X}}:=[\![ \mathbf{A},\mathbf{B},\mathbf{C}]\!] \in \mathbb{R}^{L\times M\times T}$ with a sum of $K$ rank-1 components, where $K$ is the tensor rank, i.e., 

\begin{equation}
    \underline{\mathbf{X}} \approx \sum_{k=1}^K \mathbf{a}_k \circ \mathbf{b}_k \circ \mathbf{c}_k, 
\end{equation}
where $\mathbf{A}=[ \mathbf{a}_1,\dots, \mathbf{a}_K] \in \mathbb{R}^{L\times K}$, $\mathbf{B}=[ \mathbf{b}_1,\dots, \mathbf{b}_K] \in \mathbb{R}^{M\times K}$, $\mathbf{C}=[ \mathbf{c}_1,\dots, \mathbf{c}_K] \in \mathbb{R}^{T\times K}$, and $\circ$ denotes the outer product.

The CPD of a tensor can be expressed in many different ways. For example, we can unfold $\mathbf{X}^{(1)} = \left( \mathbf{C} \odot \mathbf{B}\right) \mathbf{A}^T$ which can be seen as forming a matrix using the mode-1 fiber of the tensor
Using 'role symmetry', the mode-2 and mode-3 matrix unfolding are given by $\mathbf{X}^{(2)} = \left( \mathbf{C} \odot \mathbf{A}\right) \mathbf{B}^T$, $\mathbf{X}^{(3)} = \left( \mathbf{B} \odot \mathbf{A}\right) \mathbf{C}^T$, respectively. With only $(L+M+T) \times K$ parameters, CPD can parsimoniously represent tensors of size $L \times M \times T$.  The CPD model is a particularly powerful tool for data analysis due to two key properties: 1) every tensor admits a CPD of finite rank, indicating that it is universal, and 2) under moderate conditions, it can also be unique, i.e., it is possible to extract the true latent factors from the synthesized $\underline{\mathbf{X}}$.

\begin{table}[h]
\caption{Notations.}
 \centering
  \resizebox{0.5\textwidth}{!}{
  \begin{tabular}{c|c  }
  \hline
   Notation & Description\\
    \hline
    $\underline{\mathbf{X}}\in \mathbb{R}^{L\times M\times T}$ & spatio-temporal tensor \\
    \hline
    $\mathbf{A}\in \mathbb{R}^{L\times K}$ & location factor matrix\\
    \hline
    $\mathbf{B}\in \mathbb{R}^{M\times K}$ & signal/feature factor matrix\\
    \hline
    $\mathbf{C}\in \mathbb{R}^{T\times K}$ & temporal factor matrix\\
    \hline
    $\underline{\mathbf{X}}^{(i)}$ & mode-$i$ unfolding of $\underline{\mathbf{X}}$\\
    \hline
    $K$ & No. of hidden components\\
    \hline
    $L$ & No. of locations\\
    \hline
    $M$ & No. of features \\
    \hline
    $T$ & No. of time steps\\
    \hline
    $\circ$ & outer product \\
    \hline
    $\odot$ & Khatri-Rao product\\
    \hline
\end{tabular}}
\label{tab:notation}
\end{table}

\section{Problem Formulation}
\label{sec:problem}
Consider a location $l\in L$ where we monitor $M$ features/signals related to the RSV evolution over time, such as the number of new RSV infections, temperatures, humidity, the number of new COVID-19 infections, etc. At time $t \in T$, the value of the $m$-th signal at location $l$ is denoted as $x_{lmt}$. Given that there are $L$ locations, $M$ features and $T$ time steps, the dataset can be described as a 3-way spatio-temporal tensor $\underline{\mathbf{X}} \in \mathbb{R}^{L\times M\times T}$, where $\underline{\mathbf{X}}(l,m,t) \coloneqq x_{lmt}$. The evolution of all features and regions from $t=1$ to $t=T$ is represented by the tensor $\underline{\mathbf{X}} \in \mathbb{R}^{L\times  M\times T}$. The goal is to estimate the RSV cases from time $t=T+1$ to $t=T+T_0$ through the prediction of the frontal slabs $\underline{\mathbf{X}}(:,:,t)$ for $T_0$ time-steps in the future, where in our case $T_0$ usually sets to $52$ weeks.  


\begin{figure*}[h!]
\centering
\subfigure[\name]{\includegraphics[width=0.95\textwidth]{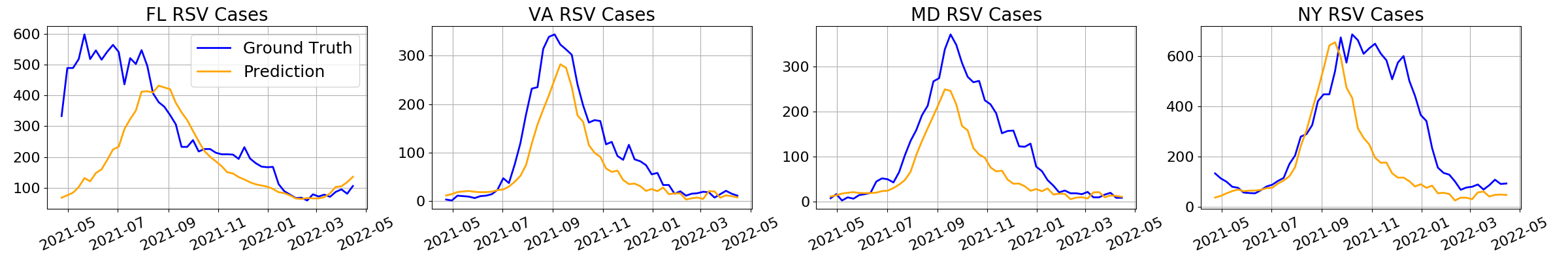}}
\vspace{-2mm}
\subfigure[DeTensor]{\includegraphics[width=0.95\textwidth]{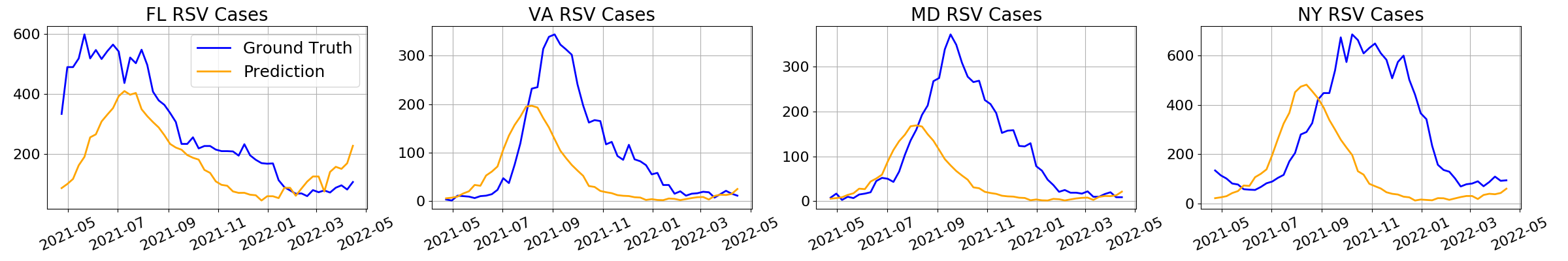}}
\vspace{-2mm}
\subfigure[LSTM]{\includegraphics[width=0.95\textwidth]{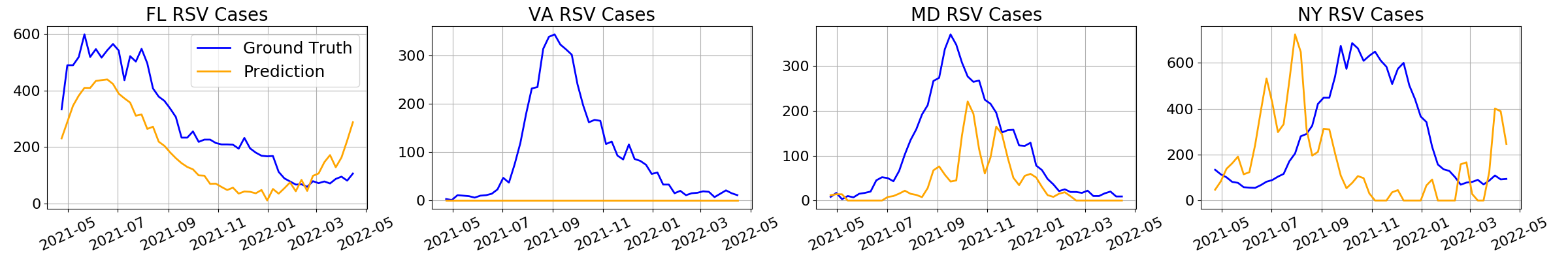}}
\vspace{-2mm}
\subfigure[ARIMA]{\includegraphics[width=0.95\textwidth]{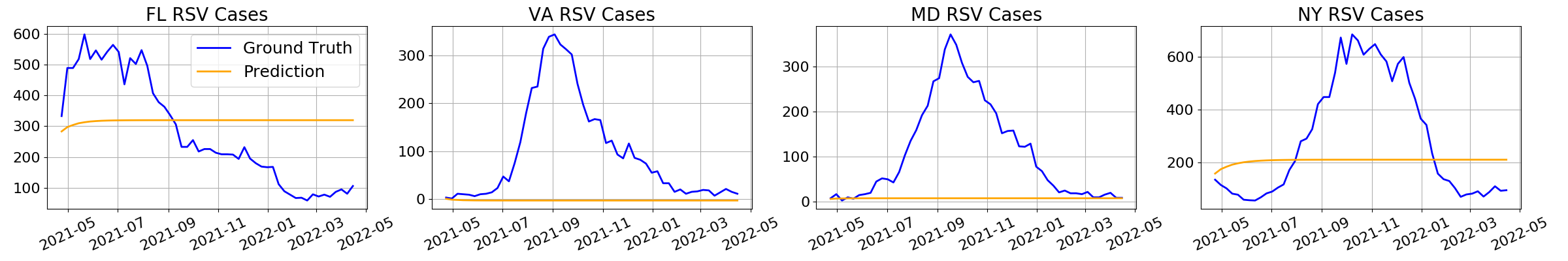}}
\vspace{-2mm}
\subfigure[SARIMA]{\includegraphics[width=0.95\textwidth]{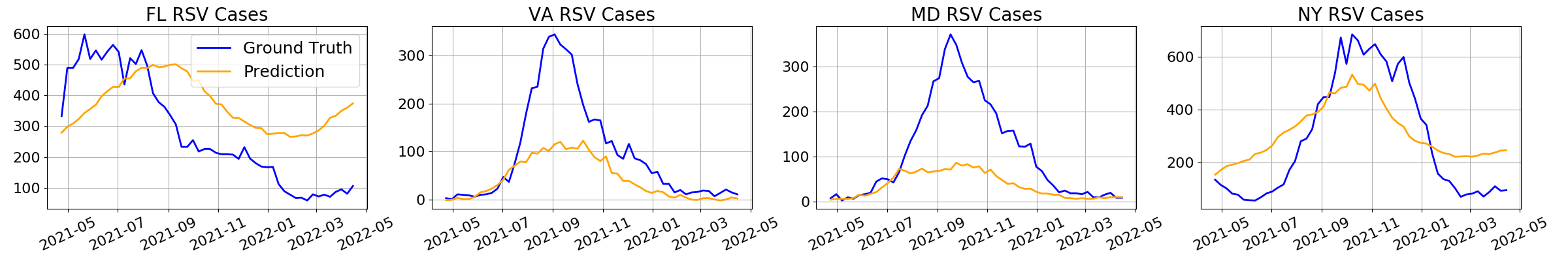}}
\vspace{-2mm}
\caption{Comparison of state-level predictive performance. 
}
\label{fig:region}
\end{figure*}

\begin{table*}[h]
\caption{State-level performance comparison in terms of RMSE and MAE of RSV cases. Results show 3 different perdition intervals: up to 12 weeks, 24 weeks, and 52 weeks.}
 \centering
  \resizebox{0.95\textwidth}{!}{
  \begin{tabular}{c|c c c c c c }
  \hline
    Model & RMSE (12 weeks) &MAE (12 weeks)& RMSE (24 weeks) & MAE (24 weeks) & RMSE (52 weeks) & MAE (52 weeks)\\
    \hline
    DeCom & \textbf{12.63(4.86)} & \textbf{11.01(3.53)}& \textbf{41.26(39.18)} & \textbf{30.64(17.35)}& \textbf{86.75(56.71)}  & \textbf{41.67(25.81)}\\
    \hline
    DeTensor  & 25.48(29.00) & 16.19(26.15) & 56.69(70.48) &32.07(49.11) & 103.09(56.74) & 55.83(34.41)\\
    \hline
    LSTM &  42.72(108.57) & 37.08(91.87) & 54.53(62.40) & 45.00(51.65) & 99.19(122.49) & 71.14(86.91)\\
    \hline
    ARIMA & 18.47(24.81) & 16.68(24.02) &78.37(92.44) & 71.20(85.05) & 100.11(115.40) & 69.34(78.36)\\
    \hline
    SARIMA & 15.91(14.26) & 14.83 (12.14) &56.24(68.17) & 50.15(61.51) & 101.76(120.21) & 71.57(85.91) \\
    \hline
\end{tabular}}
\label{tab:eval_state}
\end{table*}

\section{Proposed Method}
\label{sec:model}

\name~is a post-COVID-19 RSV prediction model with NPI awareness. We are motivated by the fact that the seasonal RSV transmission pattern was interrupted as soon as country-specific NPIs were implemented in 2020. Later on, RSV resurfaces again with the reopening happening once the number of COVID-19 cases falls below a certain level, signaling the end of the NPI control period. Therefore, the NPI control has a significant impact on breaking the normal RSV seasonality.


Figure~\ref{fig:decom} shows a high-level illustration of \name. As one can see, \name~requires two spatiotemporal tensors as input, each with three attributes: location $l$, feature $m$, and time $t$. The first tensor, denoted as $\underline{\mathbf{X}}_1 \in \mathbb{R}^{L\times M_1\times T_1} $, is the pre-COVID tensor, which includes all feature values that occurred before March 15, 2020, when the seasonal RSV transmission has not been affected by COVID-19 yet. The second tensor $\underline{\mathbf{X}}_2 \in \mathbb{R}^{L\times M_2\times T-T_1} $ denotes the post-COVID tensor, which covers the time horizon between $T_1$ and $T$.

\name~first performs a nonnegative CPD on $\underline{\mathbf{X}}_1$ to learn the subspaces $\mathbf{A}_1\in\mathbb{R}^{L\times K_1}$, $\mathbf{B}_1\in\mathbb{R}^{M_1\times K_1}$, and $\mathbf{C}_1\in\mathbb{R}^{T_1\times K_1}$ for the typical pre-COVID RSV seasons, which represent the latent subspaces of location, feature and time, respectively. Here $K_1$ denotes the rank of $\underline{\bf X}_1$. This is done by solving the following problem:
\begin{equation}
\begin{split}
\min_{\mathbf{A}_1,\mathbf{B}_1,\mathbf{C}_1} & ||\underline{\mathbf{X}}_1 - [\![ \mathbf{A}_1,\mathbf{B}_1,\mathbf{C}_1]\!] ||_F^2\\
\text{s.t.} &~\mathbf{A}_1\geq \mathbf{0},~\mathbf{B}_1\geq \mathbf{0},~\mathbf{C}_1\geq \mathbf{0}.
\end{split}
\end{equation}


Then it employs the first long short-term memory (LSTM) \citep{lstm} network $f_1(\cdot)$ to capture the normal RSV seasonality from the temporal subspace $\mathbf{C}_1$, such that the future change of $\mathbf{C}_1$ from $t=T_1$ to $t=T$ can be estimated as 
\begin{align}
    \Tilde{\mathbf{C}}_1 = f_1(\mathbf{C}_1)
\end{align}
By following the regular seasonality (i.e., without COVID-19), the RSV cases $\underline{\Tilde{\mathbf{X}}}_1$ from $t=T_1$ to $t=T$ can be approximated as 
\begin{equation}\label{X1}
    \underline{\Tilde{\mathbf{X}}}_1 = [\![ \mathbf{A}_1,\mathbf{B}_1,\Tilde{\mathbf{C}}_1]\!]  = \sum_{k=1}^{K_1} \mathbf{a}_{k,1} \circ \mathbf{b}_{k,1} \circ \Tilde{\mathbf{c}}_{k,1}
\end{equation}
where $\mathbf{a}_{k,1}$, $\mathbf{b}_{k,1}$ and $\tilde{\mathbf{c}}_{k,1}$ denote the $k$-th column of $\mathbf{A}_1$, $\mathbf{B}_1$ and $\tilde{\mathbf{C}}_1$, respectively.

The effect of COVID-19 NPIs on RSV transmission is usually reflected in the number of RSV confirmed cases. One straightforward way is to calculate the difference in RSV cases between a normal RSV season and a season with a COVID-19 impact. Note that during the COVID-19 outbreak, what we observed is the number of RSV cases under COVID-19 impact, while that number by assuming no COVID-19 can only be approximately estimated using \eqref{X1}. Therefore, we first use $\underline{\Tilde{\mathbf{X}}}_1$ to generate a residual tensor 
\begin{equation}\label{X2}
    \Delta \underline{\mathbf{X}}_2 =\underline{\mathbf{X}}_2- \underline{\Tilde{\mathbf{X}}}_1
\end{equation}
which approximates the perturbations caused by COVID-19 NPIs on a normal RSV transmission. Then we implement the second CPD on $\Delta \underline{\mathbf{X}}_2$ to learn the residual subspaces $ \mathbf{A}_2\in\mathbb{R}^{L\times K_2}$, $\mathbf{B}_2\in\mathbb{R}^{M_2\times K_2}$ and $\mathbf{C}_2\in\mathbb{R}^{(T-T_1)\times K_2}$, where $\mathbf{C}_2$ captures how COVID-19 NPIs affect RSV transmission in the latent subspace and $K_2$ is the tensor rank of $\Delta\underline{\mathbf{X}}_2$. We train the second LSTM network $f_2(\cdot)$ on $\mathbf{C}_2$ to learn the residual transmission pattern in the latent space $\mathbf{C}_2$. 

Since the last row of $\mathbf{C}_2$ corresponding to time $T$, the residual transmission pattern from $(T+1)$ to $(T+T_0)$ is predicted by passing $\mathbf{C}_2$ into $f_2(\cdot)$, i.e.,
\begin{align}
    \tilde{\mathbf{C}}_2 = f_2(\mathbf{C}_2).
\end{align}
As a result, the residual tensor from $(T+1)$ to $(T+T_0)$ is calculated as
\begin{align}
    \Delta \underline{\Tilde{\mathbf{X}}}_2 = [\![ \mathbf{A}_2,\mathbf{B}_2,f_2(\mathbf{C}_2)]\!]
     = \sum_{k=1}^{K_2} \mathbf{a}_{k,2} \circ \mathbf{b}_{k,2} \circ \Tilde{\mathbf{c}}_{k,2}.
\end{align}


\noindent 
It follows from \eqref{X1} and \eqref{X2} that the RSV cases from $(T+1)$ to $(T+T_0)$ is predicted through

\begin{equation}
\begin{split}
     \hat{\underline{\mathbf{X}}}_2 \approx 
     [\![ \mathbf{A}_1,\mathbf{B}_1,f_1(\tilde{\mathbf{C}}_1) ]\!] + [\![ \mathbf{A}_2,\mathbf{B}_2,f_2(\mathbf{C}_2)]\!]
\end{split}
\end{equation}
where $[\![ \mathbf{A}_1,\mathbf{B}_1,f_1(\tilde{\mathbf{C}}_1) ]\!]$ denotes the prediction of RSV cases without COVID-19 impact.


\begin{remark}
The advantages of our method are two-fold: 
1) It allows us to incorporate data from other sources to improve predictive power.
2) It can easily be generalized to predict other respiratory disease outbreaks under COVID-19 impact, such as influenza.

\end{remark}

\section{Experimental Setup}
\label{sec:setup}

\textbf{Data.} The dataset used in the experiments is a US state-level RSV dataset that contains weekly RSV cases for 52 US states, including Puerto Rico and Washington, D.C., between 2015 to 2022. It consists of three data sources: 1) RSV features from a large-scale medical claims database, where we extracted the state-level RSV confirmed cases and RSV related ICD-10 diagnosis features such as B79.4 (RSV), J21.0 (acute bronchiolitis due to RSV), J20.5 (acute bronchitis due to RSV), and J21.1 (pneumonia due to RSV) \citep{world2004international}. 2) COVID-19 related features from Johns Hopkins University \citep{dong2020interactive}. 3) Climate features from the Atmospheric Research Reanalysis Dataset include humidity, precipitation, and temperature \citep{reanalysis}. These features have been found to be significantly associated with the mean timing of the onset of the RSV epidemic \citep{baker2019epidemic}. 

\subsection{Evaluation pipeline.} The training data ranges from 2015-11-06 to 2021-04-16, whereas the testing data ranges from 2021-04-23 to 2022-04-15. It is worth mentioning that there was no RSV pandemic in 2020, despite testing data indicated an out-of-season RSV outbreak across all states, with peak levels exceeding prior maxima in most states. The purpose of utilizing such a training-testing split is to see if an algorithm can learn the NPI influence on RSV transmission and forecast the time and strength of the next RSV wave without having observed an RSV surge in the previous year. The experimental design does, in fact, provide a highly difficult setting for all machine learning algorithms. This way, we are able to verify if \name's NPI awareness mechanism can properly predict RSV outbreaks under the effect of COVID-19.



 \begin{figure*}[h]
\centering
\subfigure[\name]{\includegraphics[width=0.32\textwidth]{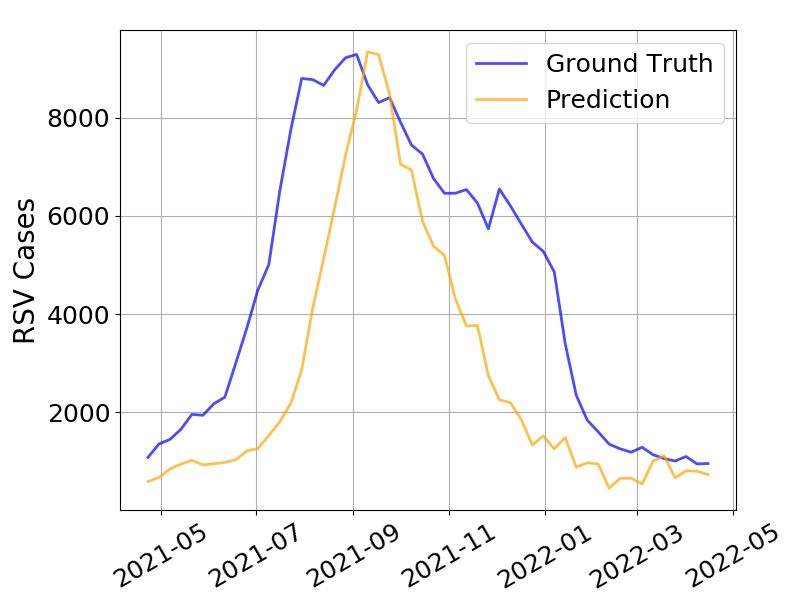}}
\subfigure[DeTensor]{\includegraphics[width=0.32\textwidth]{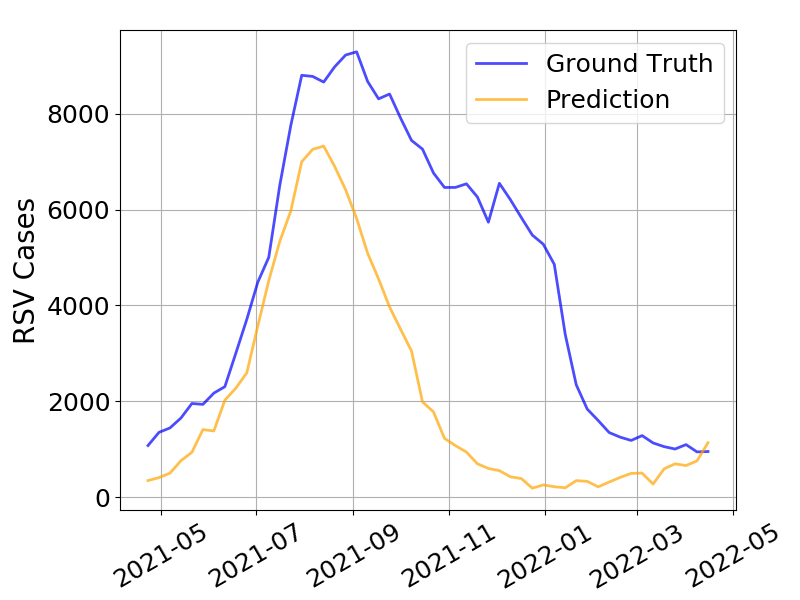}}
\subfigure[LSTM]{\includegraphics[width=0.32\textwidth]{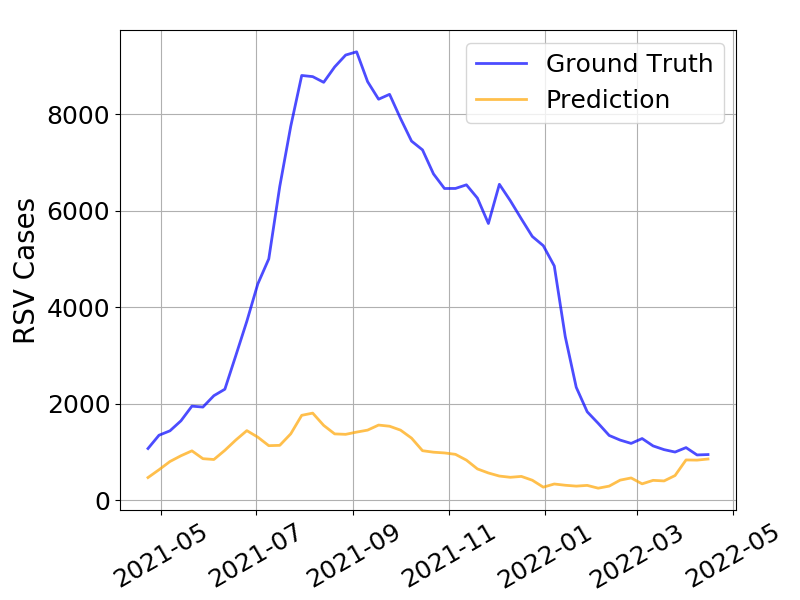}}
\subfigure[ARIMA]{\includegraphics[width=0.32\textwidth]{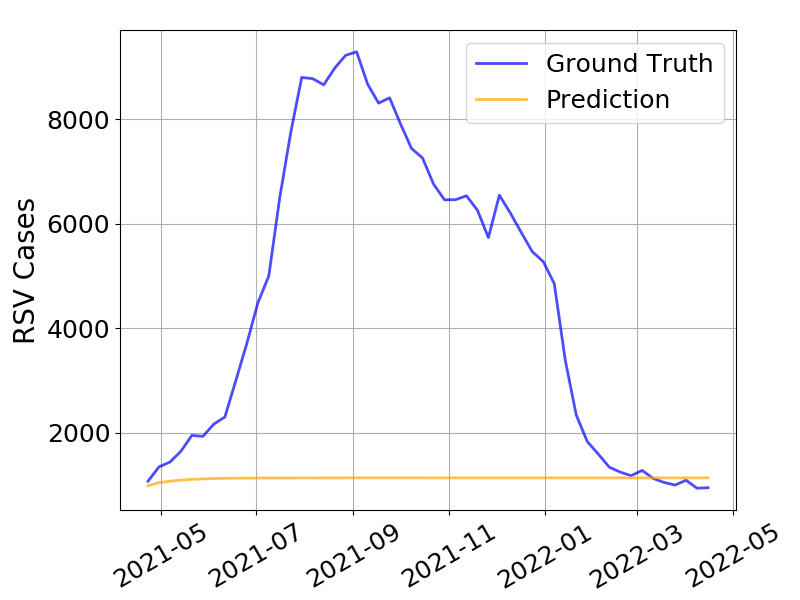}}
\subfigure[SARIMA]{\includegraphics[width=0.32\textwidth]{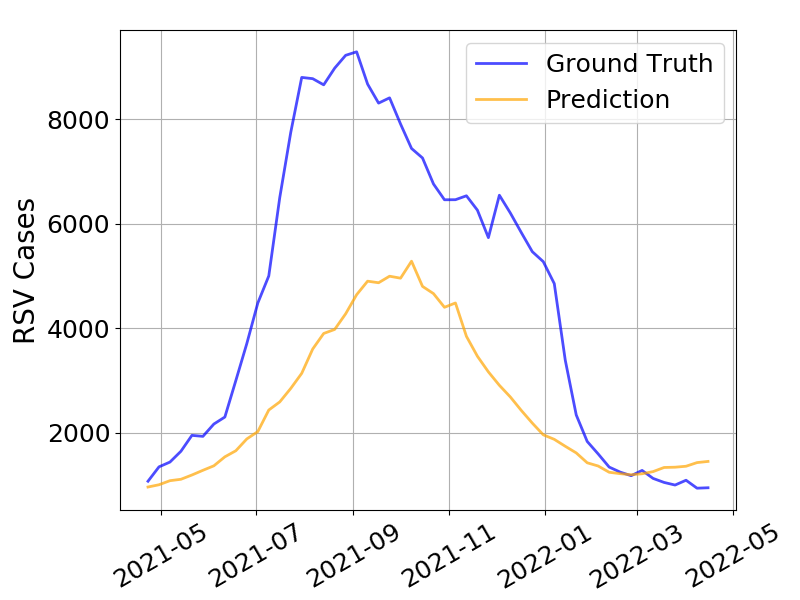}}
\caption{Country-level predictive performance for (a) \name, (b) DeTensor, (c) LSTMs, (d) ARIMA, and (e) SARIMA. \name~can accurately predict the peak time ($\pm 1$ week) and generate forecasts with significantly larger values in magnitude.
}
\label{fig:over_all}
\end{figure*}

\subsection{Evaluation metrics and baselines.} 
We report the predictive performance on the US state-level and country-level RSV data, where the latter is aggregated from 52 states, to evaluate the performance of our model. We focus on estimating RSV peak dates and weekly confirmed cases. We compute the root mean squared error (RMSE) and mean absolute error (MAE) as the performance metric.

\begin{equation}
    \label{eq:mse}
    \mathrm{RMSE} = \sqrt{\frac{1}{T}\sum_t (X_t-\hat{X}_t)^2}
\end{equation}
\begin{equation}
    \label{eq:mae}
    \mathrm{MAE} = \frac{1}{T}\sum_t |X_t-\hat{X}_t|
\end{equation}

We compare the predictive performance of \name~with deep tensor factorization model (DeTensor) \citep{kargas2021stelar}, LSTM \citep{lstm}, ARIMA, and Seasonal ARIMA (SARIMA) \citep{hyndman2018forecasting} models trained for each state. 

\subsection*{DeTensor.} 
Tensor factorization model without residual connection. Unlike proposed \name~model, DeTensor performs nonnegative CPD only once to learn the latent subspace representing a mixture of both normal RSV seasonality and the influence due to COVID-19. By comparing DeTensor, we are able to assess the effectiveness of using the second tensor in our model for NPI awareness.

\subsection*{LSTM.} The LSTM is a special Recurrent Neural Network (RNN), and it is capable of learning long-term dependencies in time series data. For every state in the dataset, we train its unique LSTM model using historical features, including COVID-19 features, for each state as input, and independently predict the RSV evolution for each state.

\subsection*{ARIMA and SARIMA.} ARIMA, or AutoRegressive Integrated Moving Average is a statistical method widely used to identify a suite of different standard temporal structures in time series data. SARIMA is an extension to ARIMA that supports the modeling of the seasonal component of the time series. The ARIMA and SARIMA models were implemented in Python using the statmodel package \cite{seabold2010statsmodels}, and the optimal parameters were found using a grid search on $(p, q, d)$. We use $s=48$ steps per circle for SARIMA.

\begin{remark}
We note that there are standard epidemiological models such as SIR-based methods for seasonal respiratory disease transmission \citep{baker2022}. The standard way is to embed a cosine function with a perios of 52 weeks to capture the seasonality of RSV. 
To implement NPI, the SIR-based approaches manually select a reduction ratio and multiply it with a pre-defined transmission rate, which may not be able to capture heterogeneities in NPIs across locations and different time windows \citep{baker2022}. Therefore, the SIR-based models that predict the arrival time of the next infectious peak strongly rely on the previously learned phase shift, the selection of transmission rate as well as when NPIs are terminated, where the latter is usually unpredictable. However, data-driven models like \name~and other deep learning models do not have such a limitation since the effect of NPIs will be reflected on the COVID-19 infections and will be learnt by the model. This is the major reason that we do not include such epidemiological methods for comparison. 
\end{remark}




\section{Experimental Results}
\label{sec:result} 

We first evaluate the state-level RSV prediction performance of \name, DeTensor, LSTM, ARIMA, and SARIMA. Figure \ref{fig:region} shows the outcomes of state-level RSV prediction, with Florida, Virginia, Maryland, and New York among the chosen states. The reason those four states have been selected is that they span from the south to the north and feature a shift in the transmission of RSV outbreaks. Therefore, we can test if an algorithm can capture the transmission difference caused by geometry differences.

Table \ref{tab:eval_state} compares the state-level prediction performance of different models over various time intervals (12, 24, and 52 weeks, respectively). We provide the mean value of RMSE and MAE as well as their standard deviation (in parenthesis). \name~has a significantly lower RMSE and MAE when focusing on the most difficult challenge of predicting RSV cases up to 52 weeks. Meanwhile, \name~outperforms other models by a wide margin in 24-week predictions and performs best for forecasts up to 12 weeks. However, the performance of DeTensor, ARIMA, and SARIMA becomes more competitive as the problem gets easier, i.e., the prediction interval gets smaller.

\begin{table}[t]
\caption{Country-level performance comparison in terms of peak time (weeks) estimates, RMSE and MAE of RSV cases. 
}
 \centering
  \begin{tabular}{c|c c c }
  \hline
    Model & Peak Diff. & RMSE & MAE\\
    \hline
    \name~ & \textbf{1} & \textbf{2461}  & \textbf{1891}\\
    \hline
    DeTensor &3 & 3175& 2485\\
    \hline
    LSTM & 4 & 4555 & 3711\\
    \hline
    ARIMA & 8 &4523& 3490\\
    \hline
    SARIMA & 5 & 2688&2102\\
    \hline
\end{tabular}
\label{tab:eval}
\end{table}

The results clearly demonstrate that, in terms of peak time and intensity estimations, \name~is superior to the other four approaches. In Maryland, for instance, the estimates of \name~are aligned with the ground truth, whereas those of DeTensor, LSTM, ARIMA, and SARIMA have shifted left or right of the true peak.


We also evaluate the effectiveness of \name~using RSV predictions at the US country-level. Specifically, we first use all five models to estimate the RSV cases in 52 US states and then accumulate the state-level cases to the country-level. The RSV estimates are plotted in Figure \ref{fig:over_all} and the performance metrics are summarized in Table \ref{tab:eval}. When the difference between the actual peak and the predicted peak is less than $\pm 1$ week, \name~can accurately predict the peak time. Furthermore, when compared to other baseline models, \name~forecasts RSV cases much closer to the ground truth, allowing it to capture the intensity more accurately. 
Overall, \name~delivers 23\% lower RMSE and 24\% lower MAE than DeTensor, 46\% lower RMSE and 49\% lower MAE compared to state-wise LSTMs, 46\% lower RMSE and MAE compared to ARIMA, and approximately $10\%$ improvement over SARIMA. 

\section{Conclusion}
\label{sec:conclusion}
In this paper, we introduce \name, a deep coupled tensor factorization machine that captures both the seasonality of RSV and the influence of COVID-19 NPIs on RSV transmission. Computer results showcase that \name~outperforms several time-series prediction baselines in predicting the timing and magnitude of RSV outbreaks under the influence of COVID-19. 
As a consequence, we believe that \name~can be an effective model to forecast and assess post COVID-19 RSV outbreaks.


\bibliographystyle{named}
\bibliography{reference}

\end{document}